\documentclass[10pt,twocolumn,letterpaper]{article}

\usepackage{cvpr}              

\usepackage{multirow}

\usepackage[sectionbib]{chapterbib}
\usepackage{amsmath,amssymb}
\usepackage{bm}
\usepackage{upgreek}

\usepackage{booktabs}
\usepackage{multirow}
\usepackage{amssymb}
\usepackage{pifont}
\usepackage{tabularx}

\usepackage{algorithm}
\usepackage{algpseudocode}

\definecolor{cvprblue}{rgb}{0.21,0.49,0.74}
\usepackage[pagebackref,breaklinks,colorlinks,allcolors=cvprblue]{hyperref}

\def\confName{CVPR}
\def\confYear{2027}

\title{Compatibility-Conditioned Policy Preservation for Continual Vision-Language-Action Learning}

\author{Li-Jun Zhao\textsuperscript{1}, Haowen Xiao\textsuperscript{1}, Haoqian Wang\textsuperscript{1}, Bairan Zhang\textsuperscript{1}, Shizhe Zhang\textsuperscript{1}\\Zhen-Duo Chen\textsuperscript{2},  Xin-Shun Xu\textsuperscript{2}, Wanhua Li\textsuperscript{1,}\thanks{Corresponding author} \\
\textsuperscript{1}Nanyang Technological University \\
\textsuperscript{2}Shandong University\\
}

\begin{document}
\maketitle
\begin{abstract}
Continual learning of vision-language-action (VLA) models aims to preserve previously learned behaviors while acquiring new manipulation tasks, requiring a careful balance between stability and plasticity. 
However, conventional preservation strategies
from continual learning, including experience
replay and knowledge distillation, face two
challenges when applied to VLA policies.
 Experience replay provides limited supervision over historical observation-instruction conditions, leaving potentially useful conditions without explicit preservation constraints. Meanwhile, knowledge distillation may introduce conflicting supervision targets when applied under the task-aligned conditions, restricting new-task acquisition. To address these challenges, we propose a novel Compatibility-Conditioned Policy Preservation (C²P²) method. Cross-condition policy preservation pairs available observations with instructions from different tasks, using distinct instruction conditions to separate preservation targets from ground-truth action supervision. Furthermore, perceptual compatibility is estimated from instruction-derived object cues and vision-language similarities to adaptively select observations and weight preservation constraints. Experiments on simulation benchmarks and real-world robotic manipulation tasks demonstrate that our method improves policy preservation and overall continual learning performance.
 Project website: \url{https://legenddddd.github.io/C2P2/}.
\end{abstract}    
\section{Introduction}
\label{sec:intro}

Continual learning is a fundamental problem
in robot policy learning, requiring robotic
policies to acquire new skills without
catastrophically forgetting previously
learned behaviors \cite{DBLP:conf/nips/LiuZGFLZS23,zheng2025imanip,liang2026robofoundry}.
In contrast to multi-task learning \cite{DBLP:conf/nips/MurugesanLCY16,DBLP:conf/icml/StandleyZCGMS20}, where
tasks are learned jointly, continual learning \cite{DBLP:journals/nn/ParisiKPKW19,de2021continual,DBLP:conf/nips/McDonnellGPAH23,DBLP:conf/icml/SzatkowskiZ0TT025}
must balance the plasticity required for
new-task acquisition with the stability
needed for knowledge retention.
Recent advances in vision-language-action (VLA) \cite{DBLP:conf/corl/ZitkovichYXXXXW23,DBLP:conf/corl/KimPKXB0RFSVKBT24,black2024pi_0}
models have provided powerful foundations for
robotic manipulation through the integration of
visual perception, language understanding, and
action generation.
Nevertheless, continual VLA learning remains
challenging due to the need to acquire new
manipulation tasks while retaining performance on previously
learned tasks.

Existing studies \cite{DBLP:journals/corr/abs-2511-18085,pmlr-v306-liu26bu,zhao2026information,hong2026fan,du2026memory} have explored established
continual learning strategies for VLA models.
However,  
continual learning for VLA models introduces distinct challenges.
Each newly introduced task provides multiple
demonstrations with a fixed language instruction,
diverse visual observations, and actions
conditioned on both.
Thus, preserving previously learned
policies requires considering the
observation-instruction conditions under
which their behaviors are generated.
In particular, experience replay (ER) \cite{chaudhry2019tiny} extends supervision to a limited set of
historical task-aligned conditions,
where observations are paired with
instructions from the same task,
to preserve previously learned behaviors.
Nevertheless, other potentially useful
cross-condition pairs combining observations and instructions
from different tasks remain outside explicit preservation
constraints, leaving room for stronger {\bf stability}.
Meanwhile, conventional knowledge distillation \cite{li2017learning}
preserves previously learned representations
or policy predictions by encouraging consistency
between the current and previous models.
However, applying it under task-aligned conditions may introduce conflicts with ground-truth action supervision in shared-head VLA policies, potentially restricting new-task {\bf plasticity}.
These limitations motivate a reconsideration
of policy preservation in continual learning of VLA models,
raising two fundamental questions:
\textit{(1) Where to preserve previous policy behaviors
beyond task-aligned conditions?}
\textit{(2) How to preserve them without directly conflicting with ground-truth action supervision?}

Answering these questions requires reconsidering
policy preservation across observation-instruction
conditions.
First, since VLA policies generate actions conditioned
on both observations and language instructions,
available observations can be paired with
instructions from different tasks to construct
additional preservation conditions.
This extends historical policy supervision
beyond the task-aligned conditions covered
by ER, enabling preservation
constraints over a broader conditional space.
Second, preserving previous policy predictions
under cross-condition historical instructions,
rather than task-aligned current instructions,
can avoid direct conflicts with ground-truth
action supervision.
Together, these insights suggest that
cross-condition policy preservation can
strengthen stability while maintaining
the plasticity required for task acquisition,
as further analyzed in
\cref{sec:conditional_dilemma}.

Building on this perspective, we propose
Compatibility-Conditioned Policy
Preservation (C$^2$P$^2$) to improve
the stability-plasticity balance in
continual learning of VLA models.
We first introduce
cross-condition policy preservation, which
broadens observation-instruction
conditions by pairing available observations with instructions from different tasks.
Under these cross-condition queries,
the previous policy provides action-chunk
pseudo-targets to preserve historical conditional
behaviors while avoiding direct conflicts with
ground-truth action supervision.
To adaptively adjust preservation strength, we
further introduce perceptual compatibility-guided
preservation.
By combining instruction-derived object
cues with vision-language similarities,
it estimates the perceptual support of
historical instructions across different
task scenes.
These estimates guide the selection of
observation sources and the weighting of
preservation constraints, enabling adaptive
policy preservation under a fixed replay
memory budget.

Our key contributions are summarized as follows:
\begin{itemize}
    
\item We delve into the stability-plasticity
dilemma in continual learning of VLA models
from a policy preservation perspective,
revealing challenges faced by conventional
preservation strategies when applied to VLA policies.

\item We propose C$^2$P$^2$, a compatibility-conditioned
policy preservation method that addresses
both where and how to preserve previous policy behaviors.

\item Extensive experiments on simulation
benchmarks and real-world robotic manipulation
tasks demonstrate benefits for both stability
and plasticity, leading to improved overall
continual learning performance.

\end{itemize}

\section{Related Work}

\noindent\textbf{Continual Learning.}
Continual learning \cite{de2021continual,zhou2024class} aims to acquire knowledge
sequentially while mitigating catastrophic forgetting,
requiring a balance between stability and plasticity.
It encompasses task-incremental, class-incremental,
and domain-incremental learning settings.
Existing approaches can be broadly categorized
into replay-based, architecture-based, and
regularization-based methods.
Replay-based methods~\cite{chaudhry2019tiny,pourkeshavarzi2021looking}
revisit historical samples or predictions to
preserve previously acquired knowledge.
Architecture-based methods \cite{yu2024boosting,DBLP:conf/iccv/WangZY25,liu2026mulora,gao2026enhanced} mitigate interference
through task-specific parameters or dynamically
allocated modules.
Regularization-based methods \cite{li2017learning,11661772,kirkpatrick2017overcoming,zhao2025attraction} constrain changes
to parameters or predictions associated with
previously learned tasks.
Among them, EWC \cite{kirkpatrick2017overcoming} mitigates forgetting by penalizing changes to parameters important for previously learned tasks, while LwF~\cite{li2017learning} distills previous model predictions using current-task data
to preserve prior knowledge during adaptation.

In task-incremental learning, task-specific output
heads can separate task supervision while sharing
underlying representations \cite{DBLP:journals/corr/abs-1904-07734}.
However, continual VLA learning typically employs
a shared action head, with task behaviors
conditioned on language instructions.
Consequently, directly applying distillation methods
such as LwF under the current instruction can
introduce conflicting targets for the shared
action head.
This motivates preserving previous policies
under appropriate instruction conditions to
mitigate such conflicts.

\noindent\textbf{Continual Learning of VLA Models.}
Vision-language-action (VLA) \cite{DBLP:conf/nips/GaoLCHFHZLFS25,DBLP:conf/cvpr/ZhangTZCCHZXXWL26} models integrate visual
perception, language understanding, and action
generation into unified robotic policies.
Continual imitation learning extends these models
to sequential task acquisition while retaining
previously learned manipulation tasks.
Representative methods, including
LOTUS~\cite{DBLP:conf/icra/WanZSZ24} and
IsCiL~\cite{DBLP:conf/nips/LeeYK0W24}, facilitate
knowledge reuse through skill organization
and efficient adaptation.

Recent studies further investigate continual
learning with pretrained VLA policies.
Continual-VLA~\cite{pmlr-v306-liu26bu,zeng2026crl} demonstrates that
pretrained VLAs exhibit strong resistance to
forgetting and benefit substantially from
limited experience replay, highlighting the
effectiveness of replay-based continual adaptation.
Stellar VLA~\cite{DBLP:journals/corr/abs-2511-18085} models
task and skill relationships through an evolving
knowledge space and employs knowledge-guided
expert routing for task specialization,
demonstrating the potential of architectural
specialization for continual VLA learning.
CLARE~\cite{DBLP:journals/ral/RomerZLS26} employs adaptive
adapter expansion and autonomous routing
to preserve previously learned behaviors.
Other works, such as AtomicVLA~\cite{DBLP:conf/cvpr/ZhangTZCCHZXXWL26}, consider a different problem setting, focusing on incremental skill expansion rather than multi-stage continual task learning.
Info-VLA~\cite{zhao2026information} focuses
on representation-level distillation rather than
the policy-level challenges specific to VLA models, 
and its evaluation employs shorter sequential task
streams.
In contrast, C$^2$P$^2$ approaches continual
VLA learning from the perspective of conditional
policy preservation, addressing the limitations
of conventional continual learning strategies
in preserving instruction-conditioned policies.


\section{Methodology}
\label{headings}

\subsection{Problem Formulation}
\label{sec:formulation}

Following prior work~\cite{pmlr-v306-liu26bu,DBLP:journals/corr/abs-2511-18085},
we consider continual VLA learning through
continual imitation learning, where a
pretrained VLA policy sequentially learns
manipulation tasks from expert demonstrations
while retaining performance on previously
learned tasks under limited access to
historical data.
Formally, a sequence of tasks
$\{\mathcal{T}_j\}_{j=0}^{J-1}$ is presented incrementally,
where each task $\mathcal{T}_j$ is specified by a language
instruction $l_j$ and a demonstration dataset
$\mathcal{D}_j=\{\tau_j^n\}_{n=1}^{N_j}$ \cite{li2024towards}.
Each demonstration $\tau_j^n$ consists of visual observations
$\mathbf{o}_t$, proprioceptive states $\mathbf{s}_t$,
and expert actions $\mathbf{a}_t$.
We define the combined observation as
$\mathbf{x}_t=(\mathbf{o}_t,\mathbf{s}_t)$
and an action chunk of horizon $H$ as
$\mathbf{A}_t=
[\mathbf{a}_t,\ldots,\mathbf{a}_{t+H-1}]^\top$.
A VLA policy
$\pi_{\bm{\uptheta}}(\mathbf{A}\mid\mathbf{x},l)$
then models the conditional distribution
of action chunks given the combined
observation and language instruction.

Under the ER protocol,
$q\%$ of the training transitions from each
completed task are retained for replay.
At stage $j$, the learner has access to
the current demonstrations $\mathcal{D}_j$
and retained transitions from previous tasks.
Let $\mathcal{X}_m^{(j)}$ denote the accessible
observations from task $m$ at stage $j$,
consisting of current-task observations when
$m=j$ and retained replay observations when
$m<j$.
At each optimization step, $B$ current-task
transitions and $B$ historical replay
transitions are randomly sampled.
For current samples
$\mathbf{x}_i\in\mathcal{X}_j^{(j)}$
and replay samples
$\mathbf{x}_r\in\mathcal{X}_{m_r}^{(j)}$
with $m_r<j$, the VLA policy is sequentially optimized using the ER objective:
\begin{equation}
\label{eq:er_objective}
\begin{aligned}
\mathcal{L}_j^{\mathrm{ER}}(\bm{\uptheta})
=\frac{1}{2B}\Bigg[
&\sum_{i=1}^{B}
\ell_{\mathrm{FM}}
(\bm{\uptheta};\mathbf{x}_i,l_j,\mathbf{A}_i)\\
+{}&
\sum_{r=1}^{B}
\ell_{\mathrm{FM}}
(\bm{\uptheta};\mathbf{x}_r,l_{m_r},\mathbf{A}_r)
\Bigg],
\end{aligned}
\end{equation}
where $\ell_{\mathrm{FM}}$ is the action flow-matching loss.

\begin{figure}[!t]
  \centering
   \includegraphics[width=1\linewidth]{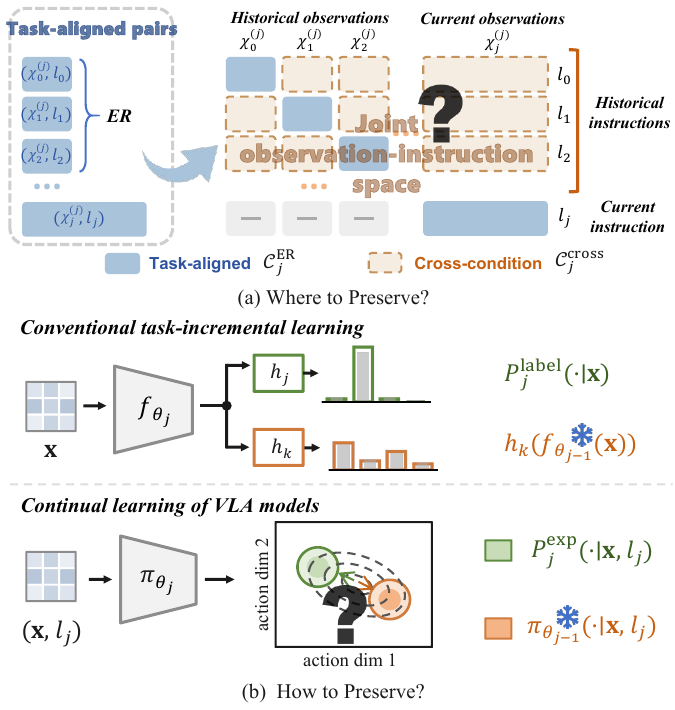}
   \caption{Illustration of the conditional preservation dilemma in continual VLA learning.
(a) 
Experience replay restricts policy preservation
to task-aligned observation-instruction conditions,
leaving cross-conditions unexploited.
(b) 
Conventional distillation under the current
instruction may introduce conflicting targets
in shared-head VLA policies, hindering new-task acquisition.
}
   \label{fig:motivation}
\end{figure}

\begin{figure*}[!t]
  \centering
   \includegraphics[width=1\linewidth]{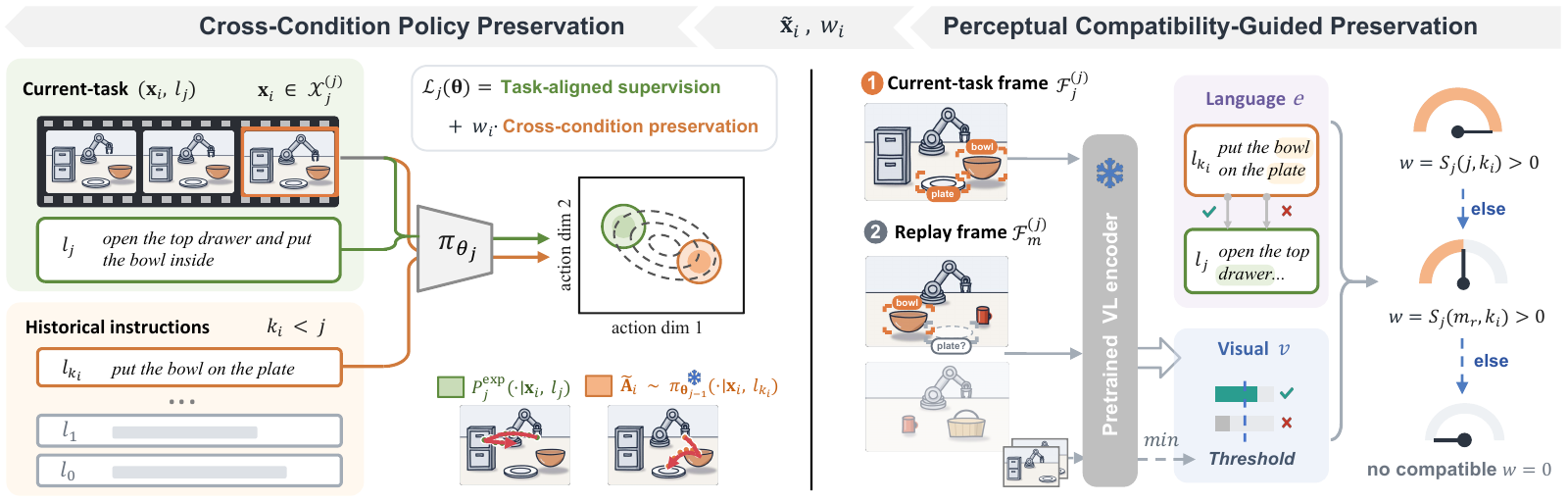}
   \caption{Overview of the proposed C$^2$P$^2$ framework.
Cross-condition policy preservation extends
policy supervision beyond task-aligned conditions
by pairing available observations with historical
instructions.
Perceptual compatibility-guided preservation
further estimates observation-instruction compatibility
to adaptively select observation sources and
weight preservation constraints.
}
   \label{fig:overall}
\end{figure*}

\subsection{The Conditional Preservation Dilemma}
\label{sec:conditional_dilemma}

Compared with conventional continual learning
settings, continual VLA learning introduces
additional challenges through instruction-conditioned action generation. Although each newly introduced task provides demonstrations associated with a fixed instruction, the policy must retain its ability to respond to previously learned instructions while acquiring new ones. Balancing the resulting stability and plasticity requirements raises two
fundamental questions in continual VLA learning.

\paragraph{Where to Preserve?}
\textit{Limited Coverage of Policy Preservation
Across Observation-Instruction Conditions.}
ER improves policy stability
by revisiting historical demonstrations.
As illustrated in \cref{fig:motivation}(a), the observation-instruction conditions covered
by the ER training distribution in
\cref{eq:er_objective} are given by
\begin{equation}
\label{eq:er_conditions}
\mathcal{C}_j^{\mathrm{ER}}
=
\bigcup_{m\leq j}
\{(\mathbf{x},l_m)
\mid\mathbf{x}\in\mathcal{X}_m^{(j)}\}.
\end{equation}
By reintroducing historical instructions,
ER extends supervision across previously learned
tasks, yet remains confined to the task-aligned
diagonal of the joint observation-instruction space.
This limited coverage leaves off-diagonal
observation-instruction conditions outside
explicit ER supervision.
Among these, conditions involving previously
learned instructions offer additional opportunities
for preserving historical policies.
These conditions are expressed as
\begin{equation}
\label{eq:cross_conditions}
\mathcal{C}_j^{\mathrm{cross}}
=
\bigcup_{k<j}
\bigcup_{\substack{m\leq j, m\neq k}}
\{(\mathbf{x},l_k)
\mid\mathbf{x}\in\mathcal{X}_m^{(j)}\}.
\end{equation}
These conditions offer broader coverage
for policy supervision, potentially strengthening
policy stability.

\paragraph{How to Preserve?}
\textit{Conflicting Supervision Targets
in Shared-Head Policy Preservation.}
Knowledge distillation provides another way
of retaining historical behaviors through
regularization.
As illustrated in \cref{fig:motivation}(b), in conventional task-incremental
learning, task-specific output heads $h_j$ and $h_k$ can separate
new-task supervision from historical distillation,
\begin{equation}
\label{eq:til_targets}
\begin{aligned}
h_j(f_{\theta_j}(\mathbf{x}))
&\rightarrow
P_j^{\mathrm{label}}(\cdot\mid\mathbf{x}),\\
h_k(f_{\theta_j}(\mathbf{x}))
&\rightarrow
h_k(f_{\theta_{j-1}}(\mathbf{x})),
\quad k<j,
\end{aligned}
\end{equation}
where $f_{\theta_j}$ and $f_{\theta_{j-1}}$
denote the shared feature extractor at the current
and previous stages, respectively, and $P_j^{\mathrm{label}}$ is the ground-truth
label distribution. 
The symbol $\rightarrow$ indicates optimization
toward the corresponding supervision target.
This separation allows new-task learning and
historical knowledge preservation to be
optimized through distinct output heads.
In contrast, continual VLA learning with a
shared action head across tasks does not
permit such separation through
task-specific output heads.
Directly applying conventional distillation-based
regularization, such as LwF~\cite{li2017learning}, under the current instruction can introduce conflicting supervision. 
Specifically, under the same observation-instruction
condition, imitation learning and distillation
optimize the shared action head toward
ground-truth actions and previous policy
predictions, respectively,
\begin{equation}
\label{eq:conditional_targets}
\begin{aligned}
\pi_{\bm{\uptheta}_j}
(\cdot\mid\mathbf{x},l_j)
&\rightarrow
P_j^{\mathrm{exp}}
(\cdot\mid\mathbf{x},l_j),\\
\pi_{\bm{\uptheta}_j}
(\cdot\mid\mathbf{x},l_j)
&\rightarrow
\pi_{\bm{\uptheta}_{j-1}}
(\cdot\mid\mathbf{x},l_j).
\end{aligned}
\end{equation}
Here, $P_j^{\mathrm{exp}}$ denotes the expert
action distribution for task $j$.
Such conflicting supervision can restrict plasticity of new behaviors, consistent with
the substantial degradation in forward transfer
observed for LwF in \cref{tab:main_results}.

\subsection{Cross-Condition Policy Preservation}
\label{sec:cross_preservation}

The preceding analysis highlights two complementary
challenges in continual VLA learning.
Native replay supervision leaves potentially useful
cross-conditions unexploited, while
align observation-instruction conditions distillation can
introduce conflicting targets that hinder task
acquisition.
Together, these challenges reveal a conditional
stability-plasticity dilemma.
To address this dilemma, we propose
\emph{Cross-Condition Policy Preservation}.
It extends preservation to additional
observation-instruction conditions using available
training data, while separating preservation targets
from new-task supervision through old instructions.

Specifically, at stage $j>0$, ER provides
supervision over the task-aligned conditions
$\mathcal{C}_j^{\mathrm{ER}}$.
To better preserve the conditional behaviors
of the previous policy, cross-condition queries
are constructed from $\mathcal{C}_j^{\mathrm{cross}}$
to extend preservation beyond these conditions.
For each cross-condition query indexed by $i$,
a historical task index $k_i$ is uniformly
sampled from $\{0,\ldots,j-1\}$.
The corresponding instruction $l_{k_i}$
is then paired with the $i$-th current-task
observation $\mathbf{x}_i\in\mathcal{X}_j^{(j)}$ in the current mini-batch,
forming a cross-condition query
$(\mathbf{x}_i,l_{k_i})$.
The frozen previous policy
$\pi_{\bm{\uptheta}_{j-1}}$ generates
an action-chunk pseudo-target under
this condition,
\begin{equation}
\label{eq:cross_target}
\widetilde{\mathbf{A}}_i
\sim
\pi_{\bm{\uptheta}_{j-1}}
(\cdot\mid\mathbf{x}_i,l_{k_i}).
\end{equation}
The generated action chunk provides
additional supervision for the current
policy through the standard flow-matching
loss.
At each optimization step, cross-condition
queries are constructed in addition to
the current and replay samples.
The resulting training objective is
\begin{equation}
\label{eq:cross_objective}
\begin{aligned}
\mathcal{L}_j(\bm{\uptheta})
={}&
\underbrace{
\mathcal{L}_j^{\mathrm{ER}}(\bm{\uptheta})
}_{\text{Task-aligned supervision}}
\\
&+
\underbrace{
\frac{1}{B}\sum_{i=1}^{B}w_i\,
\ell_{\mathrm{FM}}\left(
\bm{\uptheta};
\mathbf{x}_i,l_{k_i},
\widetilde{\mathbf{A}}_i
\right)
}_{\text{Cross-condition preservation}}.
\end{aligned}
\end{equation}
Here, $w_i\in[0,1]$ controls the preservation
strength of each cross-condition query,
with $w_i=1$ in the basic formulation.
The next subsection extends this formulation
through perceptual compatibility-guided
observation selection and adaptive weighting.

Consequently, querying the previous policy under
historical instructions $l_{k_i}$ separates
preservation targets from expert supervision
under the current instruction $l_j$,
thereby avoiding direct target conflicts.
Cross-condition preservation broadens
conditional policy supervision to strengthen
historical knowledge retention while potentially
facilitating new-task acquisition.

\subsection{Perceptual Compatibility-Guided Preservation}
\label{sec:compatibility}

Cross-condition preservation extends behavioral
supervision beyond task-aligned conditions.
However, different cross-condition queries
may contribute differently to
policy preservation due to varying degrees
of perceptual compatibility.
Therefore, we introduce
\emph{Perceptual Compatibility-Guided Preservation},
which exploits the correspondence between task
instructions and observations to adaptively
select preservation conditions and adjust their
strengths.

Specifically, to estimate perceptual compatibility,
we extract a set of entities $\mathcal{E}_m$
from each task instruction $l_m$
using linguistic rules.
For each task $m$, let $\mathcal{F}_m^{(j)}$
denote the visual frames extracted from
the accessible observations $\mathcal{X}_m^{(j)}$.
Given a set of frames $\mathcal{F}$,
the top-$K$ image-text similarities for
entity $e$ are aggregated as
\begin{equation}
\label{eq:visual_score}
g_e(\mathcal{F})
=
\frac{1}{K}\sum_{t=1}^{K}s_{(t)}(e,\mathcal{F}),
\end{equation}
where $s_{(t)}(e,\mathcal{F})$ denotes the $t$-th
highest similarity between a frame in $\mathcal{F}$
and the textual description of $e$,
computed using a frozen pretrained vision-language
encoder.
The visual evidence for entity $e$ from
task $m$ is estimated as
\begin{equation}
\label{eq:visual_evidence}
v_j(e,m)=
\begin{cases}
\mathbb{E}_{\mathcal{S}\sim\mathcal{Q}_j}
[g_e(\mathcal{S})], & m=j,\\
g_e(\mathcal{F}_m^{(j)}), & m<j,
\end{cases}
\end{equation}
where $\mathcal{Q}_j$ uniformly samples
subsets of $\mathcal{F}_j^{(j)}$ with sizes
comparable to the replay frame sets $\mathcal{F}_m^{(j)}$, and the expectation is approximated through
repeated sampling.
Combining language-derived entities
with visual evidence, the perceptual support
of entity $e$ by task $m$ is estimated as
\begin{equation}
\label{eq:entity_presence}
\widehat z_j(e,m)=
\mathbb{I}\left[
e\in\mathcal{E}_m
\;\lor\;
v_j(e,m)\geq
\min_{a\in\mathcal{A}_j(e)}v_j(e,a)
\right],
\end{equation}
where $\mathcal{A}_j(e)
=\{m\leq j\mid e\in\mathcal{E}_m\}$
denotes the set of reference tasks whose
instructions contain entity $e$,
providing an adaptive similarity threshold.
The perceptual compatibility between
observation source $m$ and instruction
$l_k$ is then defined as
\begin{equation}
\label{eq:perceptual_support}
S_j(m,k)
=
\frac{1}{|\mathcal{E}_k|}
\sum_{e\in\mathcal{E}_k}
\widehat z_j(e,m)\in[0,1].
\end{equation}
The compatibility scores are computed
once per stage and remain fixed during
policy optimization.

Finally, the estimated compatibility scores
$S_j(m,k)$ guide observation selection and
preservation weighting for each cross-condition
query.
For each sampled historical instruction $l_{k_i}$,
the current-task observation $\mathbf{x}_i$
is retained when $S_j(j,k_i)>0$.
Otherwise, a compatible observation is sought
from the replay mini-batch.
Let $\mathcal{R}_i$ denote the indices of
eligible replay samples, where each sample
$r$ originates from task $m_r\neq k_i$
and satisfies $S_j(m_r,k_i)>0$.
When $\mathcal{R}_i$ is nonempty, a replay
sample $r$ is selected with probability
\begin{equation}
\label{eq:compatibility_routing}
p(r\mid k_i)
=
\frac{S_j(m_r,k_i)}
{\sum_{r'\in\mathcal{R}_i}
S_j(m_{r'},k_i)}.
\end{equation}
Let $\widetilde{\mathbf{x}}_i$ denote the
selected observation and $m_i$ its source
task.
The selected observation replaces
$\mathbf{x}_i$ in
\cref{eq:cross_target,eq:cross_objective},
with preservation weight
$w_i=S_j(m_i,k_i)$.
If no compatible observation is available,
$w_i=0$ and the query contributes no
preservation loss.


\section{Experiments}
\label{sec:exp}

\begin{table*}[t]
\centering
\caption{Performance on the LIBERO benchmark.
The best results
are highlighted in bold.
}
\label{tab:main_results}
\setlength{\tabcolsep}{1.2pt}

\begin{tabular}{lcccccccccccc}
\toprule
\multicolumn{1}{c}{\multirow{2}{*}{\textbf{Method}}}
& \multicolumn{4}{c}{\textbf{LIBERO-Goal}}
& \multicolumn{4}{c}{\textbf{LIBERO-Long}}
& \multicolumn{4}{c}{\textbf{LIBERO-30$^*$}} \\
\cmidrule(lr){2-5}
\cmidrule(lr){6-9}
\cmidrule(lr){10-13}
& FWT $\uparrow$ & NBT $\downarrow$
& AUC $\uparrow$ & Final SR $\uparrow$
& FWT $\uparrow$ & NBT $\downarrow$
& AUC $\uparrow$ & Final SR $\uparrow$
& FWT $\uparrow$ & NBT $\downarrow$
& AUC $\uparrow$ & Final SR $\uparrow$ \\
\midrule


$\pi_0$ \cite{black2024pi_0}
& 85.0 & 74.6 & 26.5 & 14.3
& 71.5 & 63.1 & 21.1 & 8.4
& 85.3 & 76.7 & 15.7 & 7.1 \\
UniVLA \cite{DBLP:journals/corr/abs-2505-06111}
& 87.5 & 65.7 & 35.2 & 20.5
& 70.3 & 31.3 & 43.6 & 17.1
& \textbf{87.9} & 57.7 & 36.5 & 29.6 \\
MoDE \cite{DBLP:conf/iclr/ReussP0L25}
& 69.7 & 23.2 & 46.8 & 41.6
& 71.1 & 46.4 & 32.9 & 19.9
& 52.9 & 11.9 & 43.0 & 40.3 \\

\midrule
ER \cite{chaudhry2019tiny}
& 80.3 & 21.1 & 60.5 & 55.3
& 80.3 & 48.4 & 40.9 & 16.1
& -- & -- & -- & -- \\

EWC \cite{kirkpatrick2017overcoming}
& 77.4 & 68.3 & 23.4 & 11.9	
& 54.9 & 50.3 & 13.4 & 4.5
& 64.5 & 55.5 & 13.4 & 7.8 \\
LwF \cite{li2017learning}
& 9.1 & \textbf{-46.9} & 44.7 & 55.0
& 6.6 & \textbf{-17.4} & 20.0 & 21.1
& 7.3 & \textbf{-50.3} & 51.9 & 54.5 \\
\midrule
SeqLoRA \cite{DBLP:conf/iclr/HuSWALWWC22}
& 80.3 & 78.8 & 26.2 & 9.5
& 62.0 & 56.0 & 16.6 & 6.0
& -- & -- & -- & -- \\

IsCiL \cite{DBLP:conf/nips/LeeYK0W24}
& 66.8 & 7.8 & 60.9 & 48.5
& 65.8 & 31.5 & 41.4 & 17.3
& -- & -- & -- & -- \\

LOTUS \cite{DBLP:conf/icra/WanZSZ24}
& 46.8 & 37.7 & 18.5 & 10.8
& 32.3 & 2.8 & 29.4 & 31.8
& -- & -- & -- & -- \\

T-Stellar \cite{DBLP:journals/corr/abs-2511-18085}
& 76.6 & 19.1 & 57.5 & 57.0
& 80.3 & 38.0 & 48.6 & 32.2
& 43.8 & 4.1 & 38.3 & 35.8 \\

TS-Stellar \cite{DBLP:journals/corr/abs-2511-18085}
& 71.8 & 25.9 & 47.5 & 47.7
& \textbf{84.7} & 40.8 & 51.4 & 39.7
& 84.8 & 37.2 & 50.8 & 41.0 \\

Continual-VLA \cite{pmlr-v306-liu26bu}
& 85.4 & 15.3 & 72.6 & 66.5
& 64.8 & 26.5 & 42.7 & 33.8
& 80.5 & 18.4 & 63.8 & 62.2 \\

\midrule

C$^2$P$^2$ (Ours)
& \textbf{87.8} & 7.4
& \textbf{81.7} & \textbf{77.7}
& 65.0 & 13.8
& \textbf{51.5} & \textbf{48.8}
& 83.4 & 13.5 & \textbf{71.1} & \textbf{66.9} \\

\bottomrule
\end{tabular}

\end{table*}

\subsection{Experimental Setup}
\label{sec:exp_setup}

The implementation follows the training framework of Continual-VLA~\cite{pmlr-v306-liu26bu}, with initialization from the publicly released
$\pi_0$ base checkpoint of OpenPI~\cite{black2024pi_0} pretrained on cross-embodiment data.
Following Stellar VLA~\cite{DBLP:journals/corr/abs-2511-18085},
we adopt replay budgets of $1\%$ for simulation
and $5\%$ for real-world experiments.
Performance is evaluated using four metrics,
including forward transfer (FWT), negative backward
transfer (NBT), area under the success rate curve
(AUC), and final success rate (Final SR).
FWT measures new-task acquisition,
reflecting plasticity, while NBT quantifies
forgetting, reflecting stability.
AUC summarizes overall performance throughout
sequential learning, and Final SR measures
final performance after all tasks have been learned.
Together, AUC and Final SR reflect the combined
effects of plasticity and stability on
continual learning performance.
Detailed experimental setups and metric definitions are provided in the supplementary material.

\subsection{Simulation Experiments}
\label{sec:Simulation}

Following prior continual VLA learning studies ~\cite{DBLP:journals/corr/abs-2511-18085}, we evaluate
C$^2$P$^2$ on the LIBERO benchmarks
\cite{DBLP:conf/nips/LiuZGFLZS23}, namely
LIBERO-Goal, LIBERO-Long, and LIBERO-30$^*$.
These benchmarks assess continual learning
across diverse goal within a shared scene,
long-horizon tasks across diverse scenes, and
extended task sequences, respectively.
As shown in \cref{tab:main_results}, we compare
against strong pretrained VLA models adapted to
continual imitation learning, including
$\pi_0$, UniVLA, and MoDE.
We also consider representative continual learning
baselines, including ER,
EWC, and LwF, as well as recent
continual robotic learning methods, including
continual imitation learning methods such as SeqLoRA, IsCiL, and LOTUS, and continual VLA learning
approaches, including T-Stellar, TS-Stellar,
and Continual-VLA.
Among these, $\pi_0$ serves as the basic baseline,
using naive sequential fine-tuning without
experience replay or additional forgetting
mitigation strategies.
We reproduce EWC, LwF, and Continual-VLA
using the same pretrained $\pi_0$ training
setup and replay protocol as ours.

It can be observed that C$^2$P$^2$ achieves the highest AUC and Final SR
on all datasets among the
compared methods. 
In particular, our method outperforms
the best Stellar VLA variant on each dataset
in Final SR by 20.7, 9.1, and 25.9 percentage
points on LIBERO-Goal, LIBERO-Long, and
LIBERO-30$^*$, respectively.
These results demonstrate the effectiveness
of C$^2$P$^2$ across different benchmark settings.
Moreover, our method consistently outperforms
Continual-VLA across all three benchmarks
under the same training setup.
Notably, LwF achieves low NBT but suffers from
substantially reduced FWT, indicating that
preserving previous policies under new-task
conditions can severely restrict plasticity.
EWC, which regularizes model parameters
rather than predictions, slightly reduces
NBT but also lowers FWT, showing limited
effectiveness in continual VLA learning.
In contrast, C$^2$P$^2$ separates preservation
targets from new-task supervision through
instruction conditioning, avoiding direct
target conflicts. This enables effective
policy preservation while maintaining strong
new-task acquisition, as reflected by its
favorable NBT and FWT. Together, these results
support the effectiveness of our method in balancing
stability and plasticity.
Additional results on more simulation benchmarks
are provided in the supplementary material.

\begin{table}[t]
\centering
\caption{Performance on the real-world experiments. The best results
are highlighted in bold.}
\label{tab:real_world_results}
\setlength{\tabcolsep}{2pt}
\begin{tabular}{lcccc}
\toprule
\multicolumn{1}{c}{\textbf{Method}} & FWT $\uparrow$ & NBT $\downarrow$ & AUC $\uparrow$ & Final SR $\uparrow$ \\
\midrule
$\pi_0$ \cite{black2024pi_0}  & \textbf{82.0} & 66.0 & 36.2 & 16.0\\
LwF  \cite{li2017learning}   & 20.0 & \textbf{-26.8} & 37.6 & 46.0\\
Continual-VLA \cite{pmlr-v306-liu26bu} & 62.0 & -12.2 & 69.9 & 78.0\\
\midrule
C$^2$P$^2$ (Ours) & 78.0 & -4.8 & \textbf{81.4} & \textbf{86.0}\\
\bottomrule
\end{tabular}
\end{table}

\begin{table*}[!t]
\centering
\caption{Component ablation on the LIBERO benchmark. The best results are highlighted in bold.}
\label{tab:component_ablation}
\begin{tabular}{ccc|cccc|cccc}
\toprule
\multirow{2}{*}{\bf CP}
& \multirow{2}{*}{\bf CW}
& \multirow{2}{*}{\bf RO}
& \multicolumn{4}{c|}{\textbf{LIBERO-Goal}}
& \multicolumn{4}{c}{\textbf{LIBERO-Long}} \\
\cmidrule(lr){4-7}
\cmidrule(lr){8-11}
& & &
FWT $\uparrow$ & NBT $\downarrow$
& AUC $\uparrow$ & Final SR $\uparrow$
& FWT $\uparrow$ & NBT $\downarrow$
& AUC $\uparrow$ & Final SR $\uparrow$ \\
\midrule

& & 
& 85.4 & 15.3 & 72.6 & 66.5
& 64.8 & 26.5 & 42.7 & 33.8 \\

\checkmark & & 
& 87.4 & 9.4 & 79.4 & \textbf{78.0}
& 63.0 & 24.6 & 43.1 & 31.8 \\

\checkmark & \checkmark & 
& \textbf{87.8} & \textbf{7.4} & \textbf{81.7} &  77.7
& \textbf{67.0} & 22.3 & 48.6 & 36.8 \\

\checkmark & \checkmark & \checkmark
& \textbf{87.8} & \textbf{7.4} & \textbf{81.7} &  77.7
& 65.0 & \textbf{13.8} & \textbf{51.5} & \textbf{48.8} \\
\bottomrule
\end{tabular}
\end{table*}

\subsection{Real-World Experiments}
\label{sec:Real}

To further evaluate the effectiveness of
C$^2$P$^2$ beyond simulation, we conduct
additional experiments on a physical robot arm,
where five manipulation tasks are learned
sequentially under a consistent continual
learning setting.
These tasks involve ``put the blue bowl on the yellow plate'',
``stack the orange cups'',
``put the red tomato in the blue bowl'',
``pour the ball from the blue mug into the blue bowl'',
and ``open the top drawer''.
We compare against the pretrained VLA model
$\pi_0$, the regularization-based continual
learning method LwF, and the publicly available
continual VLA learning method Continual-VLA.

As shown in
\cref{tab:real_world_results}, C$^2$P$^2$ achieves the highest
AUC of 81.4\% and Final SR of 86.0\%,
consistent with its strong overall performance
in simulation.
Compared with Continual-VLA, C$^2$P$^2$
improves FWT, AUC, and Final SR, while
maintaining a negative NBT.
These improvements indicate that
our method can benefit
both new-task acquisition and historical
knowledge retention in real-world manipulation.
Although LwF achieves the lowest NBT of
$-26.8\%$, its substantially lower FWT
indicates severely restricted new-task plasticity.
In contrast, C$^2$P$^2$ maintains relatively
high FWT while achieving negative NBT,
demonstrating a more favorable balance between
stability and plasticity.

\subsection{Ablation Study}

To demonstrate the effectiveness of each component in our method,
we conduct ablation studies as reported in
\cref{tab:component_ablation}.

We first introduce cross-condition policy preservation
({\bf CP}), as described in \cref{sec:cross_preservation}.
On LIBERO-Goal, where tasks exhibit greater relatedness,
{\bf CP} reduces NBT while improving FWT, AUC, and
Final SR, indicating benefits for both stability and
plasticity and leading to better overall performance.
Notably, {\bf CP} and LwF use the same
regularization budget and training configuration,
differing only in whether policy preservation
is applied under cross-condition or task-aligned
instructions.
Compared with LwF, the substantially higher
FWT of {\bf CP} demonstrates the benefit
of separating preservation targets from
new-task supervision.
On LIBERO-Goal, {\bf CP} alone performs well,
as current-task observations provide
sufficient compatible conditions.
On the more heterogeneous LIBERO-Long,
incorporating perceptual compatibility weighting
({\bf CW}), as described in
\cref{sec:compatibility}, further improves
FWT and reduces NBT.
These improvements suggest that compatibility-guided
preservation benefits both new-task acquisition
and knowledge retention by adaptively adjusting
cross-condition preservation strength.
Finally, incorporating compatible replay observations
({\bf RO}) further improves overall continual learning
performance on LIBERO-Long by selecting compatible
observations from the replay memory.
On LIBERO-Goal, {\bf RO} brings no further improvement,
as current-task observations already provide
sufficient perceptual support for historical
instructions.

\subsection{Further Analysis}

\noindent\textbf{Different Stability-Plasticity Trade-offs.}
To further evaluate the effectiveness of
C$^2$P$^2$, we consider baseline configurations
with different stability-plasticity trade-offs
induced by learning rate schedules.

As shown
in \cref{tab:lr_ablation}, C$^2$P$^2$
improves knowledge retention and overall performance across different learning rate configurations,
indicating its effectiveness under different
initial stability-plasticity trade-offs.
Following common practice in continual learning,
task-wise learning rate restarts are adopted to
maintain plasticity, resulting in higher FWT for
the ER baseline. C$^2$P$^2$ complements this
plasticity-oriented strategy by strengthening
the preservation of previously learned tasks
while maintaining new-task acquisition.
Consequently, it reduces NBT and improves Final SR
on both benchmarks, demonstrating a better
stability-plasticity balance in continual VLA learning.

\begin{figure}[!t]
  \centering
   \includegraphics[width=1\linewidth]{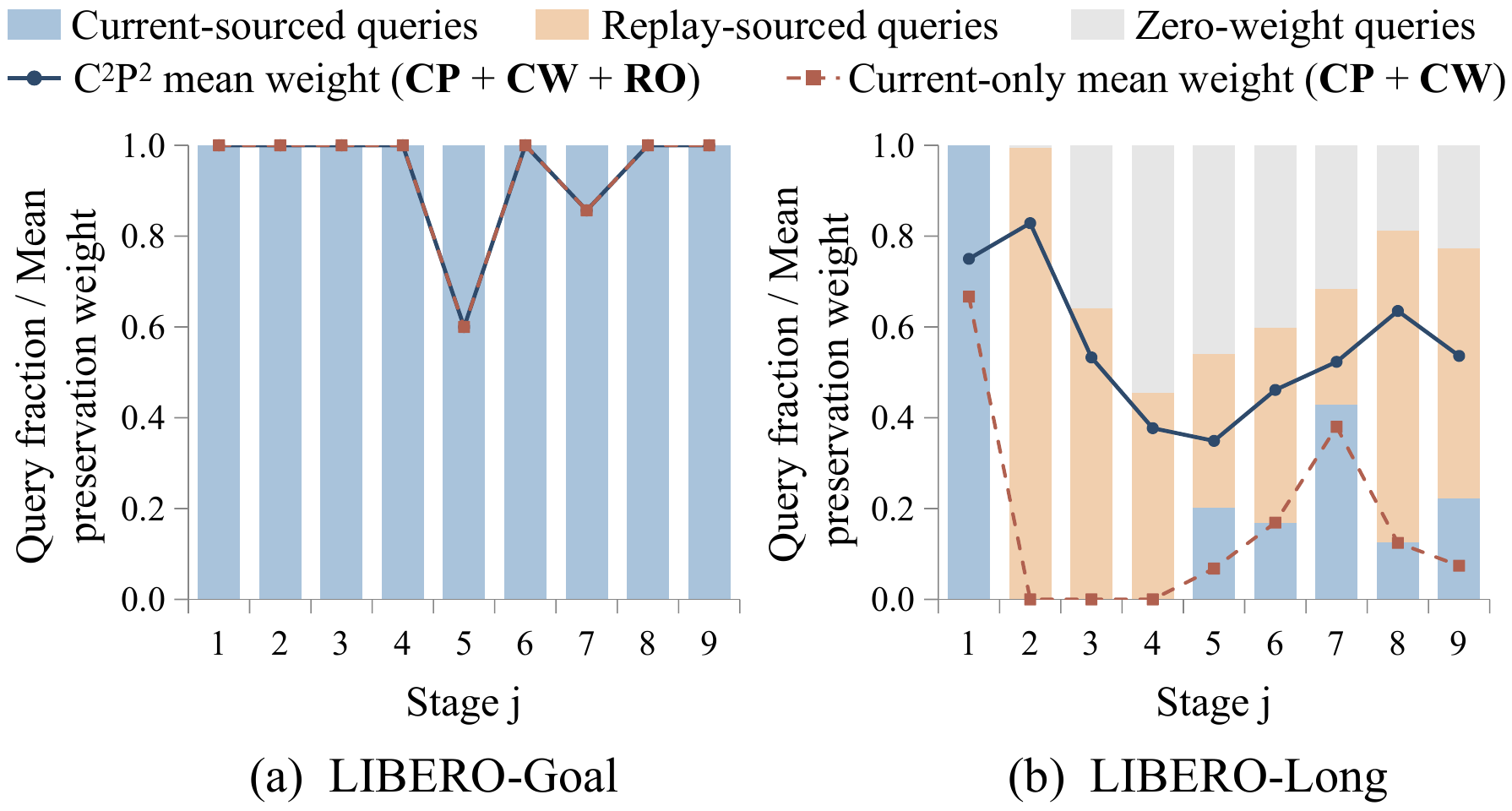}
   \caption{Adaptive policy preservation
across LIBERO-Goal and LIBERO-Long.
Query composition and mean preservation weights
illustrate how C$^2$P$^2$ adapts to heterogeneous
perceptual support.}
   \label{fig:adaptive_preservation}
\end{figure}

\begin{table*}[!t]
\centering
\caption{Effectiveness of C$^2$P$^2$ under different optimization schedules
on the LIBERO benchmark.
The best results are highlighted in bold.}
\label{tab:lr_ablation}
\begin{tabular}{llcccccccc}
\toprule
\multirow{2}{*}{\textbf{LR Schedule}}
& \multirow{2}{*}{\textbf{Method}}
& \multicolumn{4}{c}{\textbf{LIBERO-Goal}}
& \multicolumn{4}{c}{\textbf{LIBERO-Long}} \\
\cmidrule(lr){3-6}
\cmidrule(lr){7-10}
& & FWT $\uparrow$ & NBT $\downarrow$
& AUC $\uparrow$ & Final SR $\uparrow$
& FWT $\uparrow$ & NBT $\downarrow$
& AUC $\uparrow$ & Final SR $\uparrow$ \\
\midrule
\multirow{2}{*}{Original}
& baseline
& 72.2 & 12.1 & 62.0 & 59.2
& 48.8 & 14.9 & 36.3 & 28.6 \\
& + C$^2$P$^2$
& 73.6 & \textbf{2.0} & 72.3 & 69.0
& 47.8 &  \textbf{10.1} & 39.6 & 38.8 \\
\midrule
\multirow{2}{*}{Restarted}
& baseline
& 85.4 & 15.3 & 72.6 & 66.5
& 64.8 & 26.5 & 42.7  & 33.8 \\
& + C$^2$P$^2$
& \textbf{87.8} & 7.4 & \textbf{81.7} &  \textbf{77.7}
& 65.0 & 13.8 & \textbf{51.5} & \textbf{48.8} \\
\bottomrule
\end{tabular}
\end{table*}

\noindent\textbf{Adaptive Policy Preservation.}
To investigate how perceptual compatibility guides
policy preservation, \cref{fig:adaptive_preservation}
visualizes the fractions and mean preservation weights of
cross-condition queries across learning stages.
The stacked bars indicate the fractions of queries
sourced from current observations, sourced from replay
observations, or assigned zero weight due to
insufficient compatibility.
The curves compare the mean preservation weights
with and without {\bf RO}, while {\bf CP} alone
assigns a uniform weight of 1.0 to all queries.

On LIBERO-Goal, most queries are supported by
current observations, resulting in consistently
high preservation weights.
In contrast, LIBERO-Long exhibits more heterogeneous
perceptual support, where C$^2$P$^2$ adaptively
selects compatible observations and adjusts
preservation strength according to their compatibility.
This enables effective use of existing observations
without imposing uniform constraints across tasks.
These trends are consistent with the component
ablation in \cref{tab:component_ablation}, where
compatibility-aware weighting and adaptive observation
selection jointly improve policy preservation
and overall performance, particularly on
LIBERO-Long.



\begin{figure}[!t]
  \centering
   \includegraphics[width=1\linewidth]{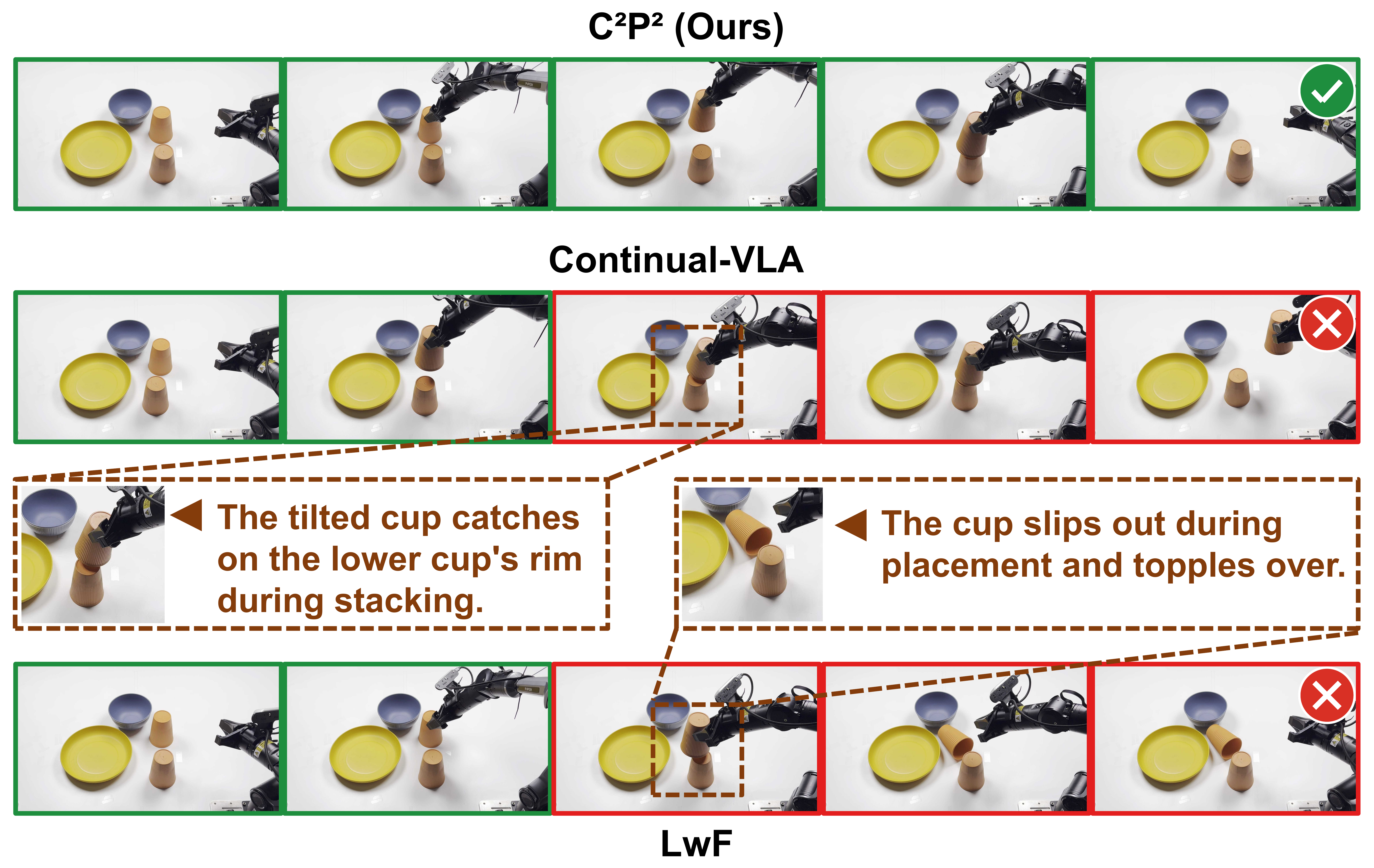}
   \caption{ 
Comparison of real-world execution trajectories
for the task ``stack the orange cups''
using models obtained after sequentially
learning all five tasks.
}
   \label{fig:task1-vis}
\end{figure}

\noindent\textbf{Qualitative Analysis of Real-World Manipulation.}
To further examine the behavioral differences
among methods in real-world manipulation,
\cref{fig:task1-vis} visualizes their execution
of the historical task ``stack the orange cups''
using the final models obtained after learning
all five tasks.
Continual-VLA exhibits action deviations
that prevent successful task completion,
such as failing to release the gripper.
Although LwF can execute the complete
action sequence, its less precise manipulation
can lead to task failure. 
In contrast, C$^2$P$^2$ maintains accurate
task execution after sequential training,
illustrating its ability to preserve
previously learned manipulation behaviors.

\begin{figure}[!t]
  \centering
   \includegraphics[width=1\linewidth]{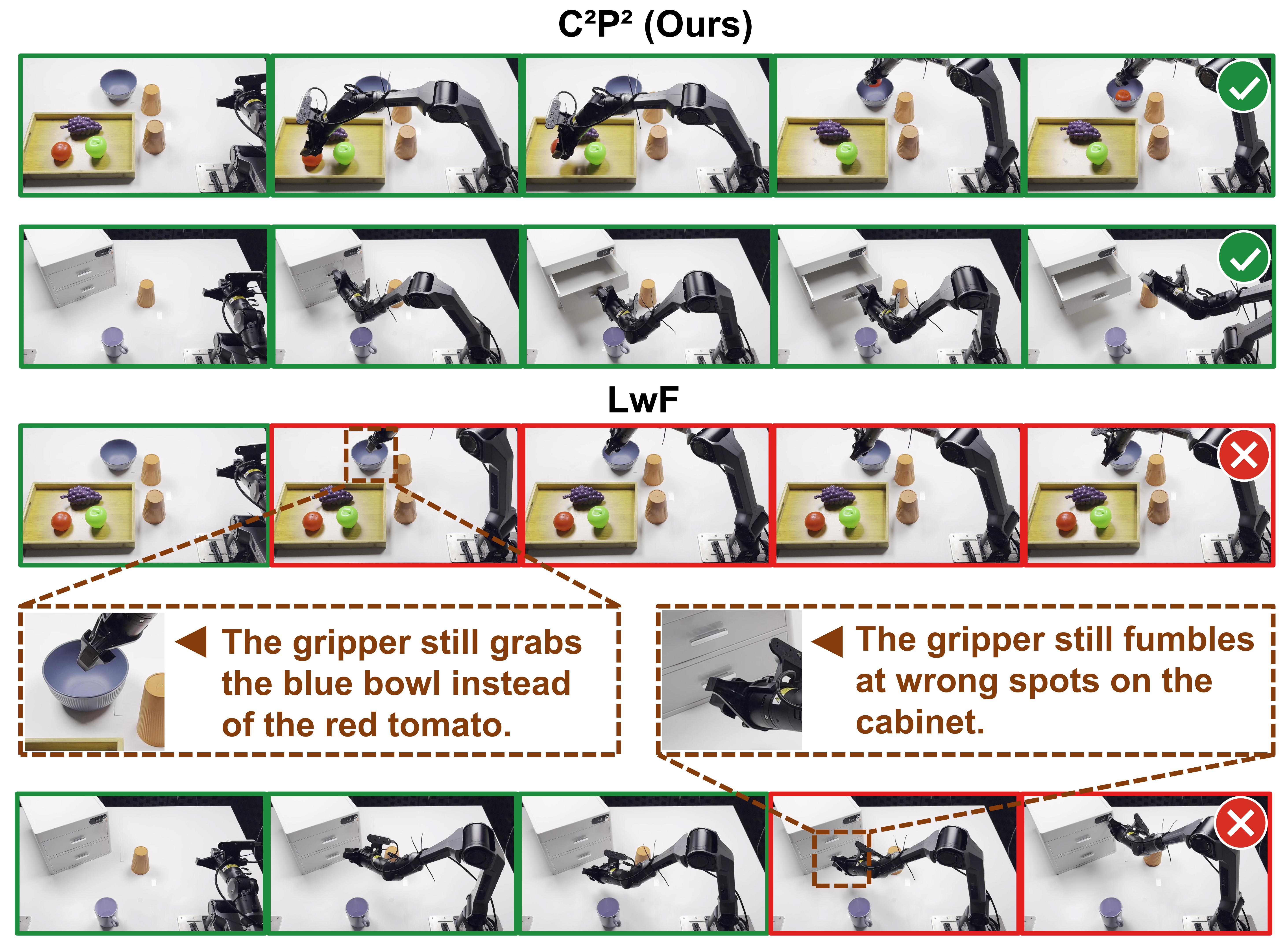}
   \caption{ 
Comparison of C$^2$P$^2$ and LwF
on the real-world tasks
``put the red tomato in the blue bowl''
and ``open the top drawer''.
Each model is evaluated immediately after
completing the corresponding training stage
to examine new-task acquisition.
}
   \label{fig:lwf-vis}
\end{figure}

We further examine new-task acquisition
in \cref{fig:lwf-vis}, which visualizes
real-world execution immediately after
the corresponding training stages.
For the task ``put the red tomato in the blue bowl'',
C$^2$P$^2$ successfully completes the manipulation,
whereas LwF incorrectly approaches the blue bowl
instead of the target tomato.
LwF also fails to open the top drawer,
repeatedly approaching incorrect locations
on the cabinet.
These observations are consistent with
the substantially lower FWT of LwF
in \cref{tab:real_world_results},
illustrating the difficulty of acquiring
new manipulation behaviors under
current-instruction distillation.

Together, these qualitative results illustrate
that C$^2$P$^2$ preserves previously learned
manipulation tasks while maintaining
strong new-task acquisition.
Additional qualitative analyses and
real-world demonstration video are provided
in the supplementary material.

\section{Conclusion}

In this paper, we investigated the conditional preservation dilemma in continual learning of VLA models. To address the challenges of conventional
preservation strategies when applied to VLA
policies, we proposed a Compatibility-Conditioned
Policy Preservation (C$^2$P$^2$) method.
Experiments on simulation benchmarks and real-world robotic manipulation tasks demonstrate that our method improves policy preservation and overall continual learning performance. 
These findings highlight the importance of accounting for the instruction-conditioned nature of VLA policies when designing continual learning methods, suggesting new opportunities for improving stability and plasticity beyond demonstration-level knowledge preservation.

\section*{LLM Usage Statement}

We used OpenAI ChatGPT and Anthropic Claude
to assist with editing and refining author-produced
text, code, and visual materials, including
language polishing, rephrasing, and grammar
correction.
These tools were also used to provide
reviewer-style feedback on the manuscript.
All research contributions, including the
methodology, experimental design, and results,
were developed by the authors.
The authors reviewed all AI-assisted content
and take full responsibility for the submitted work.

{
    \small
    \bibliographystyle{ieeenat_fullname}
    \bibliography{main}
}

\clearpage
\setcounter{page}{1}
\maketitlesupplementary

\section{Procedure for C²P²}
\label{sec:alg}

To illustrate how C$^2$P$^2$ is applied
across continual learning stages,
we summarize the complete training procedure
in \cref{alg:c2p2}.

\begin{algorithm}[h]
\caption{Training procedure of C$^2$P$^2$}
\label{alg:c2p2}
\begin{algorithmic}[1]
\Require Current stage $j>0$,
current task demonstrations $\mathcal{D}_j$,
historical replay memory $\mathcal{M}_{<j}$,
task instructions $\{l_m\}_{m\leq j}$,
previous policy $\pi_{\bm{\uptheta}_{j-1}}$
\Ensure Updated policy $\pi_{\bm{\uptheta}_j}$

\State Initialize
$\bm{\uptheta}_j\gets\bm{\uptheta}_{j-1}$
and freeze $\pi_{\bm{\uptheta}_{j-1}}$

\Statex \textit{$\triangleright$ Perceptual compatibility estimation}
\State Extract $\mathcal{E}_j$ from instruction $l_j$
and reuse historical entity sets
$\{\mathcal{E}_m\}_{m<j}$
\State Compute visual evidence $v_j(e,m)$
from image-text similarities
using Eqs.~\eqref{eq:visual_score}
and~\eqref{eq:visual_evidence}
\State Update reference task sets
$\mathcal{A}_j(e)$ 
\State Combine $\mathcal{E}_m$, $\mathcal{A}_j(e)$,
and $v_j(e,m)$ to estimate 
$\widehat z_j(e,m)$
using Eq.~\eqref{eq:entity_presence}
\State Aggregate $\{\widehat z_j(e,m)\}_{e\in\mathcal{E}_k}$
into compatibility scores $S_j(m,k)$
using Eq.~\eqref{eq:perceptual_support}

\Statex \textit{$\triangleright$ Adaptive cross-condition preservation}

\For{each training iteration}
    \State Sample $B$ current and $B$ replay
    transitions

    \For{$i=1,\ldots,B$}
        \State Sample
        $l_{k_i}\sim
        \mathcal{U}(\{l_0,\ldots,l_{j-1}\})$

        \State Set
        $\widetilde{\mathbf{x}}_i
        \gets\mathbf{x}_i$,
        $m_i\gets j$,
        $w_i\gets S_j(j,k_i)$
        \If{$w_i=0$}
            \State Form $\mathcal{R}_i$ with
            $m_r\ne k_i$ and $S_j(m_r,k_i)>0$

            \If{$\mathcal{R}_i\ne\emptyset$}
                \State Sample $r$ using
                Eq. \eqref{eq:compatibility_routing}
                \State Use replay observation
                $\widetilde{\mathbf{x}}_i
                \gets\mathbf{x}_r$
                with source task $m_i\gets m_r$
                \State Update weight
                $w_i\gets S_j(m_i,k_i)$
            \EndIf
        \EndIf

        \If{$w_i>0$}
            \State Generate pseudo-target
            $\widetilde{\mathbf{A}}_i$
            using Eq.~\eqref{eq:cross_target}
        \EndIf
    \EndFor

    \State Compute $\mathcal{L}_j(\bm{\uptheta})$
    using Eq.~\eqref{eq:cross_objective}
    \State Update $\bm{\uptheta}_j$
\EndFor
\end{algorithmic}
\end{algorithm}

\section{Detailed Experimental Setups}
\label{sec:detset}

\subsection{Evaluation Metrics Details}
\label{sec:metrics_details}

We evaluate continual VLA learning using four
metrics, including forward transfer (FWT),
negative backward transfer (NBT), area under
the success rate curve (AUC), and final success rate (Final SR).
Let $R_{j,k}$ denote the success rate on task $k$
after completing training at stage $j$, where
$0\leq k\leq j<J$ and $J$ is the total number
of tasks.

\noindent\textbf{Forward transfer (FWT).}
FWT evaluates the ability to acquire each new
task by averaging its success rate immediately
after the corresponding training stage,
\begin{equation}
\mathrm{FWT}
=
\frac{1}{J}
\sum_{k=0}^{J-1} R_{k,k}.
\end{equation}

\noindent\textbf{Negative backward transfer (NBT).}
NBT quantifies the degradation of previously
learned tasks as subsequent tasks are acquired.
For each task, the performance decrease relative
to its initial acquisition is averaged over
all later stages,
\begin{equation}
\mathrm{NBT}
=
\frac{1}{J}
\sum_{k=0}^{J-2}
\frac{1}{J-1-k}
\sum_{j=k+1}^{J-1}
\left(R_{k,k}-R_{j,k}\right).
\end{equation}
The final task contributes no degradation term.
Lower NBT indicates better knowledge retention,
while negative values indicate that subsequent
learning improves earlier task performance
on average.

\noindent\textbf{Area under the success rate curve (AUC).}
AUC measures overall performance throughout
the learning sequence by averaging each task's
success rates across all stages following
its introduction,
\begin{equation}
\mathrm{AUC}
=
\frac{1}{J}
\sum_{k=0}^{J-1}
\frac{1}{J-k}
\sum_{j=k}^{J-1} R_{j,k}.
\end{equation}

\noindent\textbf{Final success rate (Final SR).}
Final SR measures the overall policy performance
after the complete task sequence by averaging
the success rates of all learned tasks,
\begin{equation}
\mathrm{Final\ SR}
=
\frac{1}{J}
\sum_{k=0}^{J-1} R_{J-1,k}.
\end{equation}

Higher FWT, AUC, and Final SR, along with
lower NBT, indicate better continual learning
performance.

\begin{table*}[t]
\centering
\caption{Task order and language instructions
of LIBERO-Goal and LIBERO-Long benchmark suites.}
\label{tab:libero_instructions}
\begin{tabular}{c c l}
\toprule
\textbf{Benchmark} & \textbf{Task Order} & \textbf{Language Instruction} \\
\midrule
\multirow{10}{*}{LIBERO-Goal}
& 0 & open the middle drawer of the cabinet \\
& 1 & put the bowl on the stove \\
& 2 & put the wine bottle on top of the cabinet \\
& 3 & open the top drawer and put the bowl inside \\
& 4 & put the bowl on top of the cabinet \\
& 5 & push the plate to the front of the stove \\
& 6 & put the cream cheese in the bowl \\
& 7 & turn on the stove \\
& 8 & put the bowl on the plate \\
& 9 & put the wine bottle on the rack \\
\midrule
\multirow{10}{*}{LIBERO-Long}
& 0 & put both the alphabet soup and the tomato sauce in the basket \\
& 1 & put both the cream cheese box and the butter in the basket \\
& 2 & turn on the stove and put the moka pot on it \\
& 3 & put the black bowl in the bottom drawer of the cabinet and close it \\
& 4 & put the white mug on the left plate and put the yellow and white mug on the right plate \\
& 5 & pick up the book and place it in the back compartment of the caddy \\
& 6 & put the white mug on the plate and put the chocolate pudding to the right of the plate \\
& 7 & put both the alphabet soup and the cream cheese box in the basket \\
& 8 & put both moka pots on the stove \\
& 9 & put the yellow and white mug in the microwave and close it \\
\bottomrule
\end{tabular}
\end{table*}

\begin{table*}[t]
\centering
\caption{Task order, scenes, and language instructions
of LIBERO-30$^*$ benchmark suite.}
\label{tab:libero30_instructions}
\begin{tabular}{c c l}
\toprule
\textbf{Task Order} & \textbf{Scene} & \textbf{Task Instruction} \\
\midrule
0  & KITCHEN SCENE10 & close the top drawer of the cabinet \\
1  & KITCHEN SCENE10 & close the top drawer of the cabinet and put the black bowl on top of it \\
2  & KITCHEN SCENE10 & put the black bowl in the top drawer of the cabinet \\
3  & KITCHEN SCENE10 & put the butter at the back in the top drawer of the cabinet and close it \\
4  & KITCHEN SCENE10 & put the butter at the front in the top drawer of the cabinet and close it \\
5  & KITCHEN SCENE10 & put the chocolate pudding in the top drawer of the cabinet and close it \\
6  & KITCHEN SCENE1 & open the bottom drawer of the cabinet \\
7  & KITCHEN SCENE1 & open the top drawer of the cabinet \\
8  & KITCHEN SCENE1 & open the top drawer of the cabinet and put the bowl in it \\
9  & KITCHEN SCENE1 & put the black bowl on the plate \\
10 & KITCHEN SCENE1 & put the black bowl on top of the cabinet \\
11 & KITCHEN SCENE2 & open the top drawer of the cabinet \\
12 & KITCHEN SCENE2 & put the black bowl at the back on the plate \\
13 & KITCHEN SCENE2 & put the black bowl at the front on the plate \\
14 & KITCHEN SCENE2 & put the middle black bowl on the plate \\
15 & KITCHEN SCENE2 & put the middle black bowl on top of the cabinet \\
16 & KITCHEN SCENE2 & stack the black bowl at the front on the black bowl in the middle \\
17 & KITCHEN SCENE2 & stack the middle black bowl on the back black bowl \\
18 & KITCHEN SCENE3 & put the frying pan on the stove \\
19 & KITCHEN SCENE3 & put the moka pot on the stove \\
20 & KITCHEN SCENE3 & turn on the stove \\
21 & KITCHEN SCENE3 & turn on the stove and put the frying pan on it \\
22 & KITCHEN SCENE4 & close the bottom drawer of the cabinet \\
23 & KITCHEN SCENE4 & close the bottom drawer of the cabinet and open the top drawer \\
24 & KITCHEN SCENE4 & put the black bowl in the bottom drawer of the cabinet \\
25 & KITCHEN SCENE4 & put the black bowl on top of the cabinet \\
26 & KITCHEN SCENE4 & put the wine bottle in the bottom drawer of the cabinet \\
27 & KITCHEN SCENE4 & put the wine bottle on the wine rack \\
28 & KITCHEN SCENE5 & close the top drawer of the cabinet \\
29 & KITCHEN SCENE5 & put the black bowl in the top drawer of the cabinet \\
\bottomrule
\end{tabular}
\end{table*}

\subsection{Simulation Experiments} 

\subsubsection{Benchmark Details}

\paragraph{LIBERO.} Following prior continual VLA learning studies~\cite{DBLP:journals/corr/abs-2511-18085}, we evaluate C$^2$P$^2$ on three benchmark suites from LIBERO \cite{DBLP:conf/nips/LiuZGFLZS23}, namely LIBERO-Goal, LIBERO-Long, and LIBERO-30$^*$.

LIBERO-Goal consists of manipulation tasks
involving different goal configurations, while
LIBERO-Long contains more complex tasks requiring
multi-step manipulation behaviors.
Each suite comprises ten tasks.
LIBERO-30$^*$ consists of the first 30 tasks
from LIBERO-90, covering multiple kitchen
scenes with some tasks sharing similar objectives
under different scene configurations.
All tasks are introduced sequentially
without joint base-task training.
Following the standard LIBERO dataset setup,
each task provides 50 expert demonstrations
for policy training.
During continual learning, the model is trained
on the current task while retaining only a
limited replay memory from previously learned
tasks.
The complete task sequences and corresponding
language instructions for LIBERO-Goal and
LIBERO-Long are provided in
\cref{tab:libero_instructions}, while those
for LIBERO-30$^*$ are provided in
\cref{tab:libero30_instructions}.

\paragraph{CALVIN.} We further evaluate C$^2$P$^2$ on a continual
learning task sequence constructed from the
CALVIN benchmark \cite{DBLP:journals/ral/MeesHRB22} using the
\texttt{task\_D\_D} split.
We select ten tasks from the official task
definitions across six categories, including
rotation, pushing, opening and closing,
lifting, light control, and pushing into a drawer.
Tasks are selected in a round-robin manner
to ensure coverage of all six categories.
The tasks are introduced sequentially without
joint base-task training.
For each task, a fixed instruction from the official
language annotations is used for each task
during both training and evaluation.
Each task is evaluated individually from
initial states satisfying its preconditions.
As on LIBERO, the model is trained on the
current task while retaining only a limited
replay memory from previously learned tasks.
The complete task sequence and corresponding
language instructions are provided in
\cref{tab:calvind_tasks}.

\begin{table*}[t]
\centering
\caption{Task order and language instructions
of our continual learning task sequence constructed from CALVIN.}
\label{tab:calvind_tasks}
\begin{tabular}{c l}
\toprule
\textbf{Task Order} & \textbf{Language Instruction} \\
\midrule
0 & take the red block and rotate it to the right \\
1 & go push the red block right \\
2 & push the sliding door to the left side \\
3 & grasp and lift the red block \\
4 & use the switch to turn on the light bulb \\
5 & slide the block that it falls into the drawer \\
6 & take the red block and rotate it to the left \\
7 & go push the red block left \\
8 & push the sliding door to the right side \\
9 & lift the red block from the sliding cabinet \\
\bottomrule
\end{tabular}
\end{table*}

\subsubsection{Experimental Details}

All compared methods are evaluated under
pretrained settings.
ER \cite{chaudhry2019tiny}, SeqLoRA \cite{DBLP:conf/iclr/HuSWALWWC22}, IsCiL \cite{DBLP:conf/nips/LeeYK0W24}, and LOTUS \cite{DBLP:conf/icra/WanZSZ24}, whose results
are reported by Stellar VLA \cite{DBLP:journals/corr/abs-2511-18085}, use LIBERO-90
pretraining, while all other compared methods
use cross-embodiment pretraining.
Our reproduced baselines, including EWC \cite{kirkpatrick2017overcoming},
LwF \cite{li2017learning}, and Continual-VLA \cite{pmlr-v306-liu26bu}, as well as
C$^2$P$^2$, are initialized from the
same cross-embodiment pretrained $\pi_0$ model.
For a controlled comparison, LwF and
the CP-only variant use the same number
of additional distillation samples and
identical training configurations.

For policy optimization, each task is trained
for 10,000 optimization steps using AdamW
with a cosine learning rate schedule.
The peak learning rate is set to
$2.5\times10^{-5}$, with a weight decay
of $10^{-10}$.
Following common practice in continual learning,
we restart the learning rate schedule with
the same peak learning rate for each task
to maintain plasticity throughout sequential learning.
Each policy is evaluated over 100 trials
with 50 initial states, and the results are
averaged over three random seeds.

For perceptual compatibility estimation
in \cref{sec:compatibility}, the entity set
$\mathcal{E}_k$ consists of object noun phrases
extracted from instruction $l_k$.
Two entity phrases are considered a match
if they share the same head noun.
Entity-level similarity scores are computed
using a frozen SigLIP encoder
\cite{DBLP:conf/iccv/ZhaiM0B23}
on third-person camera images.
For each entity $e$, the score of a frame set
is computed as the mean of the top-5 image-text
similarity logits using the text prompt
``a photo of a $\{e\}$''.
Since all current-task frames are available,
$\mathcal{Q}_j$ samples 100 frames without
replacement from $\mathcal{F}_j^{(j)}$
to match the replay scale.
The current-task score is averaged over
1{,}000 such samples using a fixed random seed.
Following the continual learning protocol,
compatibility is estimated once per stage
using only current demonstrations,
previously encountered instructions,
and retained replay samples.

\begin{table}[t]
\centering
\caption{Overall performance on CALVIN benchmark. The best results
are highlighted in bold.}
\label{tab:calvind_results}
\setlength{\tabcolsep}{2pt}
\begin{tabular}{lcccc}
\toprule
\textbf{Method} & \textbf{FWT} $\uparrow$
& \textbf{NBT} $\downarrow$
& \textbf{AUC} $\uparrow$
& \textbf{Final SR} $\uparrow$ \\
\midrule
$\pi_0$ \cite{black2024pi_0}  & 74.0 & 54.7 & 30.5 & 17.9 \\
LwF \cite{li2017learning}     & 26.2 & \textbf{-29.6} & 47.3 & 54.8 \\
Continual-VLA \cite{pmlr-v306-liu26bu}       & 76.2 & 8.2 & 69.7 & 66.6 \\
\midrule
C$^2$P$^2$ (Ours)
           & \textbf{76.5} & 2.0 & \textbf{74.9} & \textbf{73.6} \\
\bottomrule
\end{tabular}
\end{table}
\subsubsection{More Results}

To further evaluate C$^2$P$^2$ across
different manipulation environments,
we conduct additional experiments on CALVIN, as shown in \cref{tab:calvind_results}.

C$^2$P$^2$ achieves the highest AUC of 74.9\%
and Final SR of 73.6\%, outperforming
Continual-VLA by 5.2 and 7.0 percentage points,
respectively.
Meanwhile, C$^2$P$^2$ reduces NBT from
8.2\% to 2.0\% while maintaining comparable FWT.
These results demonstrate that the benefits
of C$^2$P$^2$ extend beyond LIBERO,
improving knowledge retention
without compromising new-task acquisition.

\subsubsection{Analysis of Perceptual Compatibility}
\label{sec:supp_instr_ablation}

To further investigate the effectiveness of
perceptual compatibility estimation, we compare
C$^2$P$^2$ with three alternative compatibility
estimation strategies, including $S\equiv0$
(all incompatible), $S\equiv1$ (all compatible),
and a language-only variant based on entities
mentioned in instructions.
Compatibility accuracy is evaluated against
scene-object sharing derived from BEHAVIOR
Domain Definition Language (BDDL)
task files~\cite{DBLP:conf/corl/Srivastava0LMXV21}.

\begin{figure}[t]
    \centering
    \includegraphics[width=\linewidth]{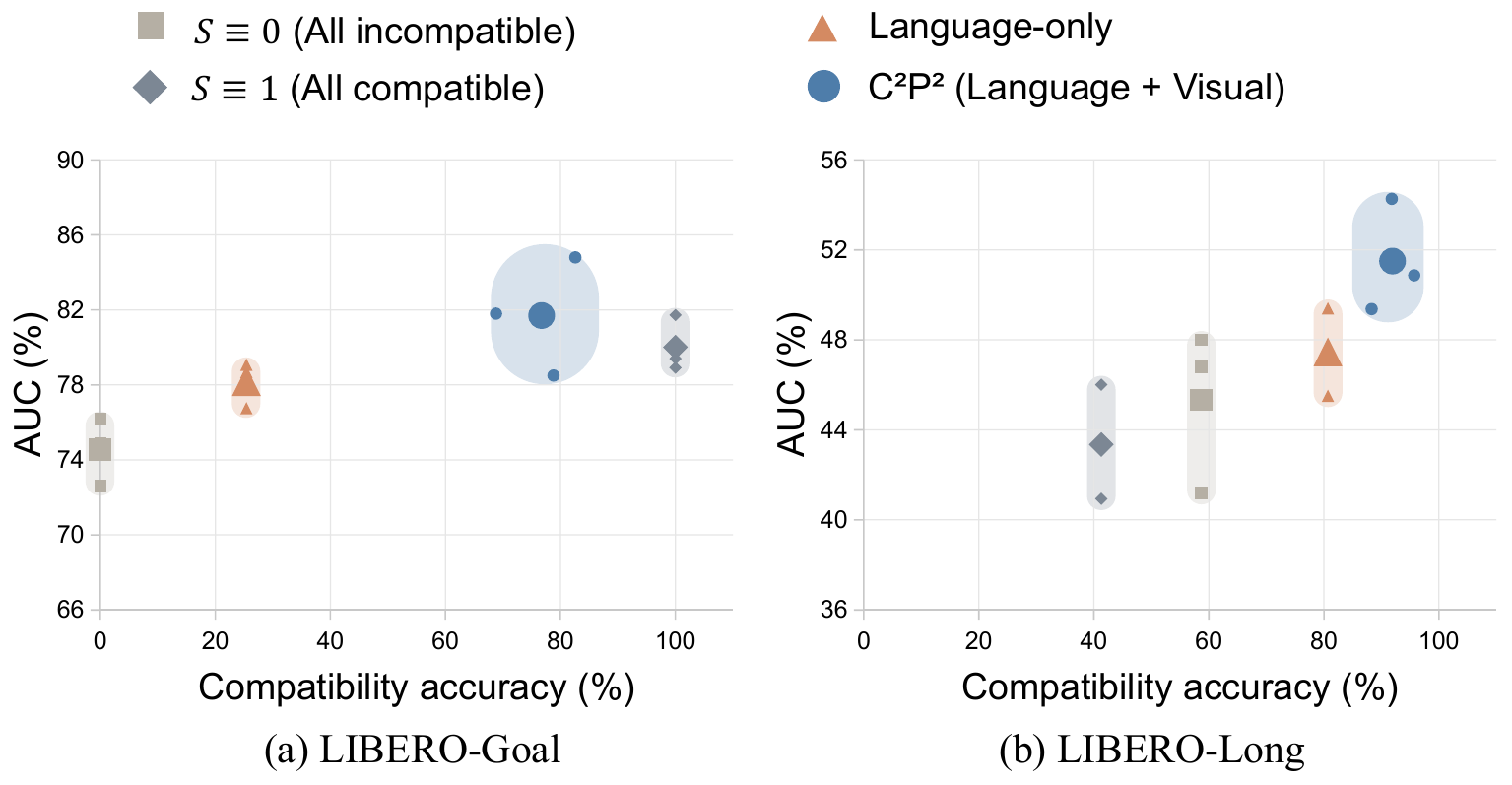}
    \caption{
    Relationship between compatibility estimation
    accuracy and continual learning performance.
    Large markers indicate mean results, small markers represent individual seeds, and shaded regions indicate the range across three seeds.
    }
    \label{fig:compatibility_ablation}
\end{figure}

As shown in \cref{fig:compatibility_ablation},
incorporating visual evidence improves
compatibility accuracy on both LIBERO-Goal
and LIBERO-Long compared with language-only
estimation.
Meanwhile, C$^2$P$^2$ achieves the highest
mean AUC on both benchmarks.
These results demonstrate that combining
language and visual evidence enables more
accurate compatibility estimation, which
contributes to improved continual learning
performance across different manipulation
settings.

\subsubsection{More Visualizations}

\begin{figure*}[t]
\centering
\begin{subfigure}{\textwidth}
    \centering
\includegraphics[width=0.9\textwidth]{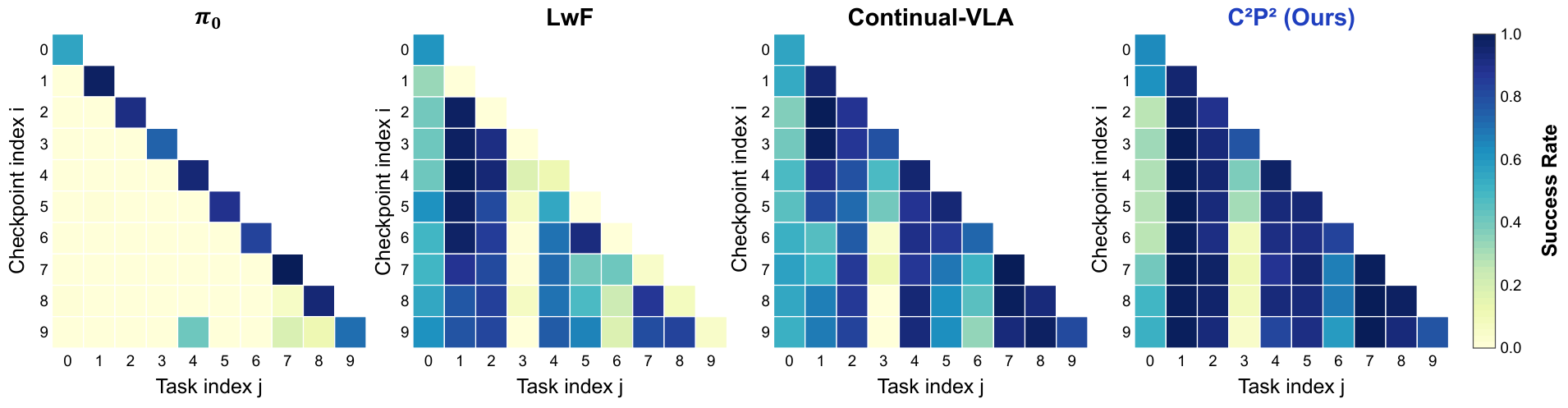}
\caption{LIBERO-Goal}
\end{subfigure}

\begin{subfigure}{\textwidth}
    \centering
\includegraphics[width=0.9\textwidth]{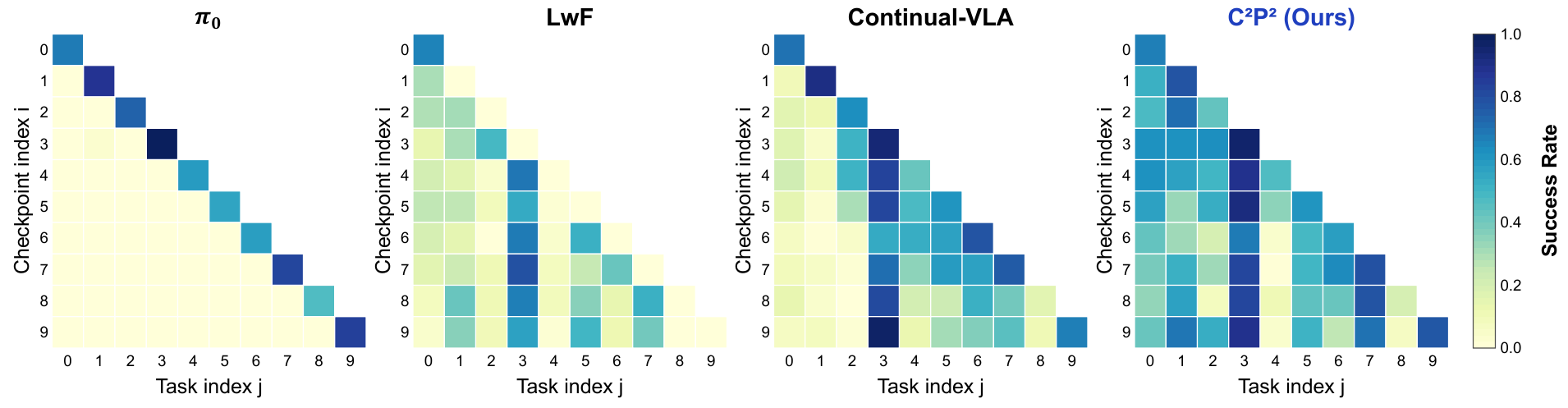}
\caption{LIBERO-Long}
\end{subfigure}

\begin{subfigure}{\textwidth}
    \centering
\includegraphics[width=0.9\textwidth]{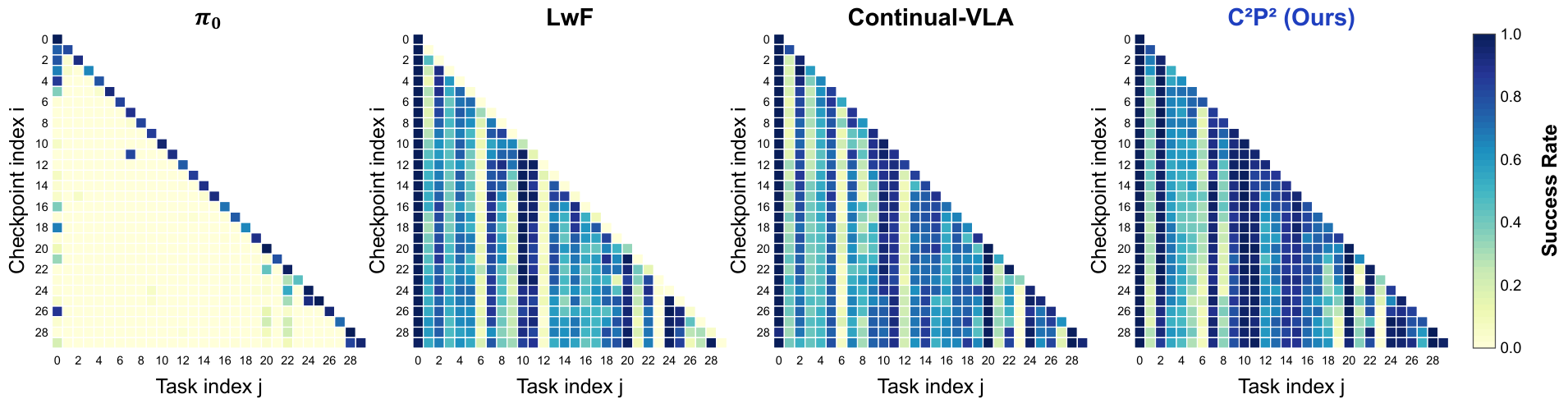}
\caption{LIBERO-30$^*$}
\end{subfigure}

\begin{subfigure}{\textwidth}
    \centering
\includegraphics[width=0.9\textwidth]{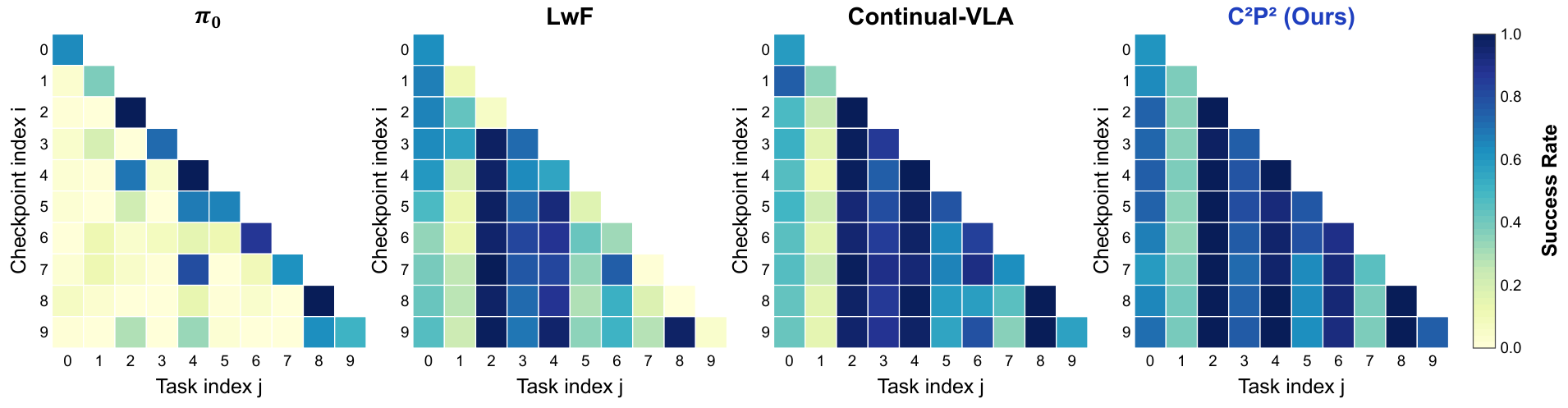}
\caption{CALVIN}
\end{subfigure}

\caption{
Task-wise success rate matrices of different
methods throughout sequential learning on
LIBERO-Goal, LIBERO-Long, LIBERO-30$^*$,
and CALVIN, respectively.
Each matrix entry $(i, j)$ denotes the success rate on Task $j$ after training on Task $i$.
Darker colors indicate higher success rates.
}
\label{fig:success_matrices}
\end{figure*}

We further visualize the task-wise success rate
matrices of different methods throughout sequential
learning on LIBERO-Goal, LIBERO-Long,
LIBERO-30$^*$, and CALVIN in
\cref{fig:success_matrices}.
Each matrix presents the success rates of
previously encountered tasks at different
learning stages, illustrating both new-task
acquisition and historical knowledge retention.

It can be observed that $\pi_0$, corresponding to naive
sequential fine-tuning, suffers from severe
forgetting.
Due to conflicting supervision on new tasks,
LwF exhibits limited new-task acquisition
and may even improve performance on previously
learned tasks at the expense of learning
new tasks.
Although Continual-VLA substantially mitigates
forgetting through experience replay,
C$^2$P$^2$ generally maintains higher success
rates across historical tasks.
These results further illustrate the effectiveness
of C$^2$P$^2$ in balancing stability and
plasticity across different continual
manipulation benchmarks.

\begin{figure*}[!t]
  \centering
   \includegraphics[width=1\linewidth]{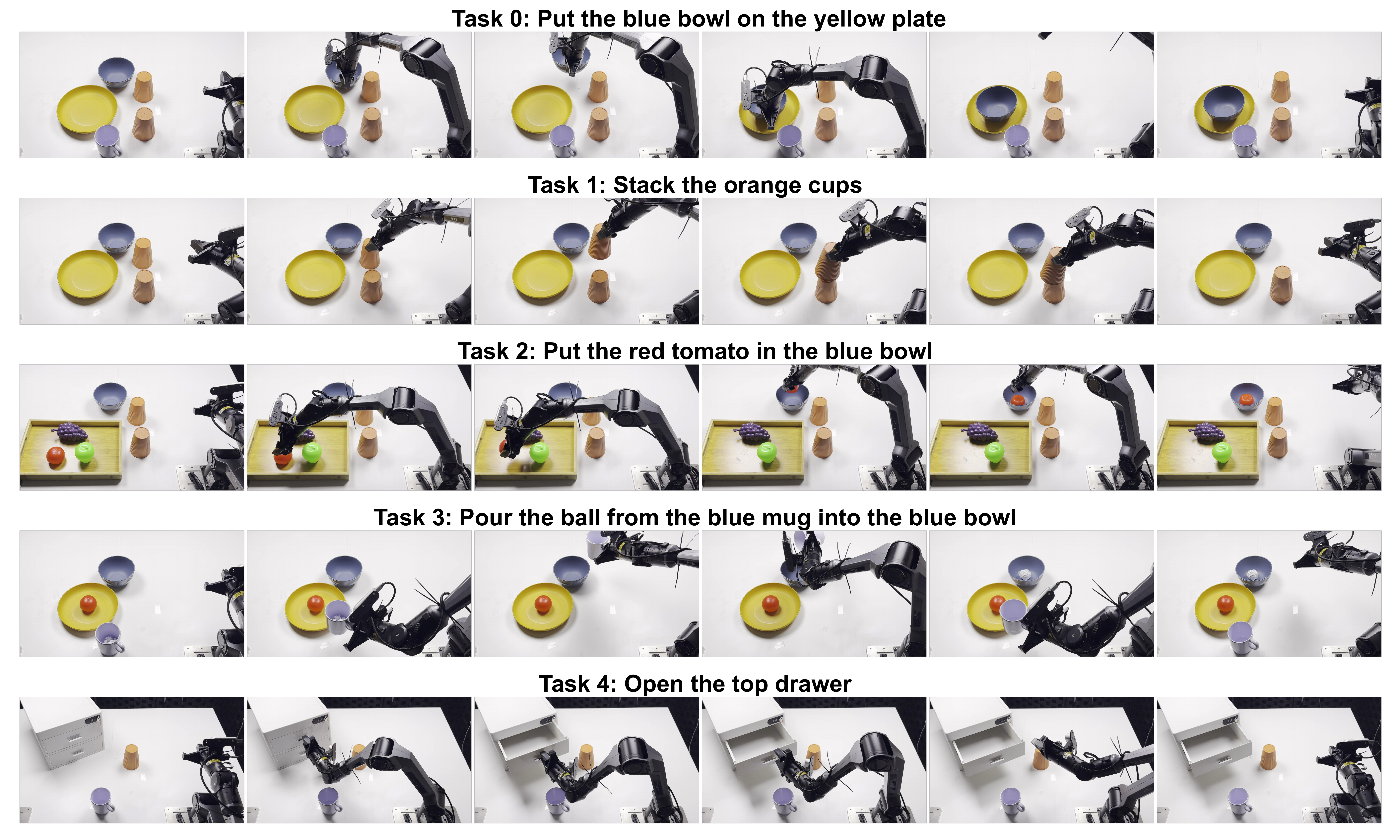}
   \caption{ Demonstration of 5 real-world tasks. We provide demonstrations of the trajectories for real-world datasets in sequential order and all real-world experiments are conducted using this sequence for continual learning training.
}
   \label{fig:real-vis}
\end{figure*}
\subsection{Real-World Experiments}
\subsubsection{Experimental Details}
\label{sec:real-robot-experiment-setup}

Our hardware platform consists of a
six-degree-of-freedom AgileX Piper
robotic arm equipped with a parallel-jaw
gripper.
Visual observations are provided by two
Intel RealSense D435 RGB cameras, one capturing
a fixed third-person view and the other
mounted on the robot's wrist.
Both cameras capture images at a resolution
of $640 \times 480$ pixels, with images resized to
$224 \times 224$ pixels for model input.
The action space is seven-dimensional,
with the first six dimensions representing
relative joint-position changes and the
last dimension specifying an absolute
gripper command.
Actions are executed at a frequency of 30 Hz.
The task stream consists of five sequential
manipulation tasks, listed in training order
in \cref{fig:real-vis}.
All methods use 50 demonstrations per task
for training and 10 evaluation trials per task.

\begin{figure*}[!t]
  \centering
   \includegraphics[width=1\linewidth]{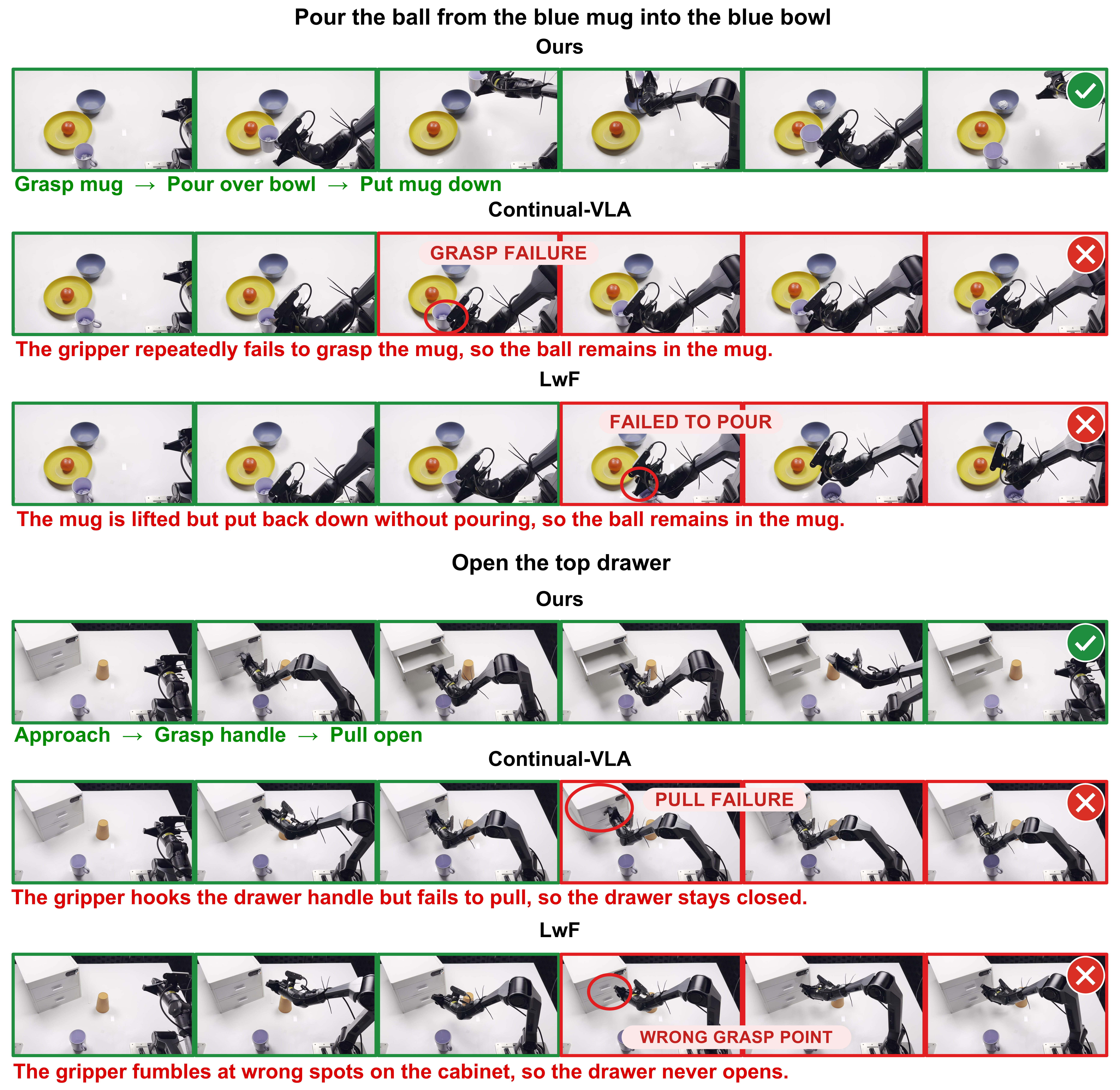}
   \caption{Additional qualitative comparisons of
real-world manipulation on the tasks
``pour the ball from the blue mug into
the blue bowl'' and ``open the top drawer''.
All methods are evaluated after completing
the full five-task training sequence.
}
   \label{fig:task34-vis}
\end{figure*}

\subsubsection{Additional Qualitative Results}

To further examine the behavioral differences
among methods, additional real-world execution
visualizations are provided in \cref{fig:task34-vis}.

Consistent with the preceding observations,
Continual-VLA exhibits difficulties in precise
manipulation, such as failing to grasp the mug
or pull open the drawer after engaging its handle.
Although LwF can complete more intermediate
steps in some tasks, such as lifting the mug,
it still fails to execute the required actions,
including pouring and correctly grasping
the drawer handle.
In contrast, C$^2$P$^2$ successfully completes
both manipulation tasks, further illustrating
its ability to preserve previously learned
policy behaviors under sequential learning.

\end{document}